\documentclass[letterpaper, 10 pt, conference]{ieeeconf}  

\IEEEoverridecommandlockouts                     
\usepackage{graphics} 
\usepackage{epsfig} %
\usepackage{mathptmx} 
\usepackage{times} 
\usepackage{amsmath} 
\usepackage{amssymb}  
\usepackage{mathrsfs}
\usepackage{graphicx}
\usepackage{caption}  
\usepackage{subcaption}
\usepackage{algorithmic}
\usepackage{algorithm}
\usepackage{hyperref}
\usepackage{graphicx}
\usepackage{bbm}
\usepackage{enumerate}
\usepackage[usenames, dvipsnames]{color}
\usepackage{makecell}

\DeclareMathAlphabet{\mathcal}{OMS}{cmsy}{m}{n}

\newcommand{\ccalH}{\mathcal{H}}

\input{mysymbol.sty}

\title{\LARGE \bf Composable Learning with Sparse Kernel Representations}
\author{ Ekaterina Tolstaya$^1$, Ethan Stump$^2$, Alec Koppel$^2$, Alejandro Ribeiro$^1$ 
\thanks{This work is supported by grants NSF DGE-1321851 and ARL DCIST CRA W911NF-17-2-0181.}
\thanks{$^{1}$Department of ESE, University of Pennsylvania, Philadelphia, PA 19104, USA  \{eig, aribeiro\}@seas.upenn.edu.}
\thanks{$^{2}$Computational and Information Sciences Directorate, U.S. Army Research Laboratory, Adelphi, MD 20783, USA \{ethan.a.stump2.civ, alec.e.koppel.civ, \}@mail.mil.}
}
\begin{document}
\maketitle
%
\begin{abstract}
We present a reinforcement learning algorithm for learning sparse non-parametric controllers in a Reproducing Kernel Hilbert Space.  
We improve the sample complexity of this approach by imposing a structure of the state-action function through a normalized advantage function (NAF). This representation of the policy enables efficiently composing multiple learned models without additional training samples or interaction with the environment.
We demonstrate the performance of this algorithm on learning obstacle-avoidance policies in multiple simulations of a robot equipped with a laser scanner while navigating in a 2D environment. We apply the composition operation to various policy combinations and test them to show that the composed policies retain the performance of their components. We also transfer the composed policy directly to a physical platform operating in an arena with obstacles in order to demonstrate a degree of generalization.
\end{abstract}
%
 
\section{Introduction}
 \begin{figure} 
\includegraphics[width=0.93\linewidth] {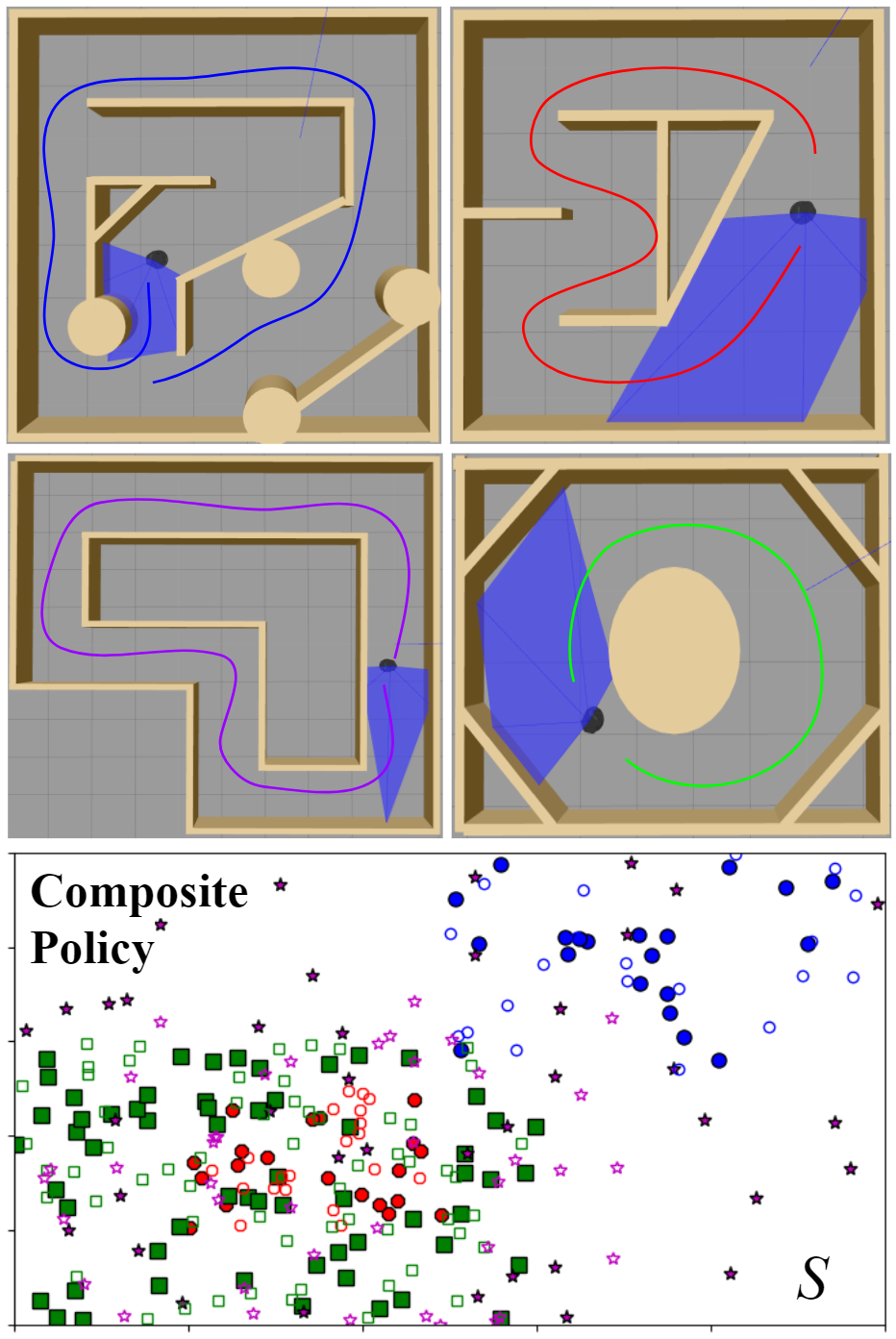}
\caption{ 
This drawing illustrates the concept of four policies, trained on individual environments and each represented with their own kernel support, being composed into a single policy with broader support that applies everywhere. These four training environments (referred to as Maze, Circuit 1, Circuit 2, and Round) were used in simulation as part of the demonstration of our approach to composable learning. 
}
\label{fig:diagram}
\end{figure}

A key goal in the design of distributed robot teams is the ability to learn collaboratively, so that knowledge and experience gained by one system can be seamlessly incorporated in another. Parallel experiences have already been used to stabilize and hasten joint learning \cite{mnih2016asynchronous}, but we argue that there is also a need for learning techniques designed for loosely-coupled teams, in which members may be disconnected for long periods of time and then briefly reconnect to share information and coordinate. We re-consider joint learning problems where coordination is now an infrequent event and not part of the update cycle. 

This problem requires \emph{composable learning}, the capability for models learned by different systems to be directly composed as a single model that combines the strengths of each component. We need model representations that can be joined without additional training, which is a major challenge for modern data-driven  (deep) machine learning. Different agents with different experiences will invariably develop different representations that cannot be directly combined. Though the agents may use the same deep neural network architecture, each structurally-equivalent weight will be specialized for different purposes.

To capture this notion of composability, we choose instead to use models that live in a Reproducing Kernel Hilbert Space (RKHS),  a classic non-parametric learning approach, in which a learned model is represented directly over its training data \cite{scholkopf2002learning}. Because the domain of the learned model's applicability is directly represented by its support, we can imagine how models that were developed over separate regions of the data space can combine their supports into one broader model. Recent work on online learning of sparse kernel functions \cite{polk} gives us tools for building larger kernel functions while minimizing the number of basis points required in the support.

We apply this model composition approach to the scenario of robots learning to avoid obstacles by experiencing collisions, a well-studied problem that has been successfully solved on physical platforms using methods that blend reinforcement learning with foundational techniques in optimal control \cite{richter2017safe} or supervised learning \cite{kahn2018self}. Because our primary interest is understanding the composability of obstacle-avoiding policies, we opt to learn our policies using Q-learning as a stand-in for more sophisticated approaches, based on recent work in Q-learning with sparse kernel representations \cite{tolstayaacc}. To enable learning control policies in the continuous action space of our problem, we propose a kernelized version of the Normalized Advantage Function Q-learning algorithm \cite{naf}.

In the context of Q-learning, the density of kernel support of the converged policy is a reflection of the states where we can expect that policy to be near optimal, since these states are drawn from the stationary distribution of the Bellman optimality operator. We suggest the use of the density of the support to decide which component policy should be chosen to dominate in a composition of policies in a region of the state space. We can represent this density directly in the RKHS as an application of the Kernel Mean Embedding \cite{muandet2016kernel}, but with the addition of our sparsity-inducing procedures.

The \textbf{main contributions} of this work are:
\begin{enumerate}
\item An algorithm for composing multiple RKHS functions according to their density of kernel support.
\item A Kernel Normalized Advantage Function Q-learning algorithm that extends \cite{naf} to learn RKHS policies.
\item A proof-of-concept demonstration in which we: (a) train obstacle-avoidance policies in multiple simulations of a robot equipped with a laser scanner and navigating in a 2D environment; (b) apply composition to various policy combinations and test them to show that the composed policies retain the performance of their components; and (c) transfer the composed policy directly to a physical platform operating in an arena with obstacles in order to elicit a sense of generalization.
\end{enumerate}


\section{Normalized Advantage Functions in RKHS}

We model an autonomous agent in a continuous space as a Markov Decision Process (MDP) with continuous states $\mathbf{s}\in\mathcal{S} \subseteq \mathbb{R}^p$ and actions $\mathbf{a}\in\mathcal{A} \subseteq \mathbb{R}^q$. When in state $\mathbf{s}$ and taking action $\mathbf{a}$, a random transition to state $\mathbf{s}'$ occurs according to the conditional probability density $\mathbb{P}(\mathbf{s}' | \mathbf{s}, \mathbf{a})$. 
After the agent to a particular $\mathbf{s}'$ from $\mathbf{s}$, the MDP assigns an instantaneous reward $r(\mathbf{s},\mathbf{a},\mathbf{s}')$, where the reward function is a map $r: \mathcal{S} \times \mathcal{A} \times \mathcal{S} \rightarrow \mathbb{R}$.  

In Markov Decision problems\cite{bellman1954theory}, the goal is to find the action sequence $\{ \mathbf{a}_t \}_{t=0}^\infty$ so as to maximize the infinite horizon accumulation of rewards, i.e.,  the value: $V(\mathbf{s},\{ \mathbf{a}_t \}_{t=0}^\infty) := \mathbb{E}_{\mathbf{s}'} \lbrack \sum_{t=0}^\infty \gamma^t r(\mathbf{s}_t,\mathbf{a}_t,\mathbf{s}'_t) \mid \mathbf{s}_0 = \mathbf{s}, \{ \mathbf{a}_t \}_{t=0}^\infty  \rbrack$ \cite{kqlearning}. The action-value function $Q(\mathbf{s},\mathbf{a})$ is the conditional mean of the value function given the initial action $\mathbf{a}_0=\mathbf{a}$:
\begin{align} \label{eq:qdef}
Q(\!\mathbf{s},\mathbf{a},\!\!\{\!\mathbf{a}_t\! \}_{t=1}^\infty\!) 
\!\! & :=  \\  \! \mathbb{E}_{\mathbf{s}'} & \!\!\!\left[\!  \sum_{t=0}^\infty \!\gamma^t \! r(\mathbf{s}_t,\mathbf{a}_t,\mathbf{s}'_t) \!\!   \mid \!\mathbf{s}_0\!\! =\! \mathbf{s}, \mathbf{a}_0\! = \!\mathbf{a}, \! \{ \!\mathbf{a}_t \!\}_{t=1}^\infty  \!\!\right]  \nonumber
\end{align}
We consider the case where actions $\mathbf{a}_t$ are chosen according to a stationary stochastic policy, where a policy is a mapping from states to actions: $\pi$: $\mathcal{S} \rightarrow \mathcal{A}$.  We define $Q^*(\mathbf{s},\mathbf{a})$ as the maximum of \eqref{eq:qdef} with respect to the action sequence. 

It's possible to formulate finding the optimal action-value function as a fixed point problem \cite{bertsekas2004stochastic}: shift the index of the summand in \eqref{eq:qdef} by one, make use of the time invariance of the Markov transition kernel, and the homogeneity of the summand, to derive the Bellman optimality equation :
\begin{equation} \label{eq:qstar} 
Q^{*}\!(\mathbf{s},\mathbf{a}) \! = \!\! \! \mathbb{E}_{\bbs'} \! \lbrack r(\mathbf{s},\mathbf{a},\mathbf{s}') +  \gamma  \max_{\mathbf{a}'} \! Q(\mathbf{s}'\!,\mathbf{a}') \rbrack  \;  
\end{equation}
The reason for defining action-value functions is that the optimal  $Q^*$ may be used to compute the optimal policy $\pi^*$ as
\begin{equation} \label{eq:pistar} 
\pi^*(\mathbf{s}) = \argmax_\mathbf{a} Q^*(\mathbf{s},\mathbf{a}) \; .
\end{equation}

Computation of the optimal policy for continuous action problems requires maximizing the $Q$ function \eqref{eq:pistar} which may not have a closed form, and can be challenging. To mitigate this issue, we hypothesize that $Q$ has a special form: it is a sum of the value and an \emph{advantage} \cite{baird1994reinforcement}:
\begin{align}\label{eq:Q_hypothesized_form}
Q(\bbs,\bba) = V(\bbs) + A(\bbs,\bba)
\end{align}

According to the definition of the action-value function, we require that $\max_\bba Q(\bbs,\bba) = V(\bbs)$. Therefore, the optimal action has no advantage: $\max_{\bba}A^*(\bbs,\bba) = 0$.

For this hypothesis to yield computational savings, we parametrize the advantage function as a quadratic function  \cite{naf}. We define $L(\bbs):\ccalS \rightarrow \ccalA \times \ccalA $ as a matrix function, so $L^T(\bbs)L(\bbs)$ is positive definite. The policy is a bounded vector function $\pi:\ccalS \rightarrow \ccalA$.  
\begin{align}
A(\bbs,\bba) = -\frac{1}{2} (\bba - \pi(\bbs)) L^T(\bbs) L(\bbs) (\bba - \pi(\bbs))
\end{align}
Thus, we assume the advantage function is quadratic in the actions, which is restrictive, albeit sufficient, for the applications we examine. 

To find the optimal policy, we seek to satisfy \eqref{eq:qstar} for all state-action pairs, yielding the cost functional $\tilde{J}(Q)$:
\begin{align} 
\tilde{J}(V,\pi,L) = \mathbb{E}_{\bbs,\bba} (y(\bbs,\bba) - Q(\bbs, \bba))^2,  
\end{align}
where $y(\bbs,\bba) = \mathbb{E}_{\bbs'}[ r(\bbs,\bba,\bbs') + \gamma V(\bbs')]$.
Substituting \eqref{eq:Q_hypothesized_form}, finding the Bellman fixed point reduces to the stochastic program:
\begin{align}  \label{eq:main_prob} 
V^{*},L^{*},\pi^* = \argmin_{V,\pi,L\in \ccalB(\ccalS)} 
 \tilde{J}(V,\pi,L) \; .
\end{align}
Since searching over value functions that are arbitrary bounded continuous functions $\ccalB(\ccalS)$ is untenable, we restrict $\mathcal{B}(\mathcal{S})$ to be a reproducing Kernel Hilbert space (RKHS) $\ccalH$ to which $V(\bbs)$ belongs \cite{kqlearning}. Further, the quadratic parameterization of the advantage function means that $\pi$ and $L$ also belong to $\ccalH$. An RKHS over $\ccalS$ is a Hilbert space is equipped with a reproducing kernel, an inner product-like map $\kappa:\mathcal{S}\times\mathcal{S} \rightarrow \mathbb{R}$ \cite{norkin2009stochastic,argyriou2009there}:
\begin{align} \label{eq:repprop}
 \text{(i)} \langle \pi, \kappa(\bbs, \cdot ) \rangle_\ccalH  =  \pi(\bbs) , \;\;\;\;\;
\text{(ii)} \ccalH =  \text{span} \lbrace \kappa(\bbs,\cdot) \rbrace 
\end{align}
In this work, we restrict the kernel used to be in the family of universal kernels, such as a Gaussian kernel with constant diagonal covariance $\Sigma$, i.e., 
\begin{align}  \label{eq:rbfdef}
\kappa(\mathbf{s},\mathbf{s}')  =  \exp \left\lbrace  - \frac{1}{2} \left(\mathbf{s}-\mathbf{s}'\right)\Sigma \left(\mathbf{s}-\mathbf{s}'\right)^T  \right\rbrace 
\end{align}
motivated by the fact that a continuous function over a compact set may be approximated uniformly by a function in a RKHS equipped with a universal kernel \cite{micchelli2006universal}. 

To apply the Representer Theorem, we require the cost to be coercive in $V$, $\pi$ and $L$ \cite{argyriou2009there}, which may be satisfied through use of a Hilbert-norm regularizer, so we define the regularized cost functional $J(V,\pi,L) = \tilde{J}(V,\pi,L) + (\lambda/2) (\|V\|_{\ccalH}^2 + \|\pi\|_{\ccalH}^2 + \|L\|_{\ccalH}^2 )$ and solve the regularized problem.
\begin{align}  \label{eq:lossprob} 
   V^{*},\pi^{*},L^{*} &= \argmin_{V,\pi,L\in \ccalH} J(V,\pi,L) \\
         &= \argmin_{V,\pi,L\in \ccalH} \tilde{J}(V,\pi,L) +  \frac{\lambda}{2} (\|V\|_{\ccalH}^2 + \|\pi\|_{\ccalH}^2 + \|L\|_{\ccalH}^2 ) . \nonumber
\end{align}
We point out that this is a step back from the original intent of solving \eqref{eq:main_prob} because the assumption we have made about $V^{*}$, $\pi^{*}$, and $L^{*}$ being representable in the RKHS $\ccalH$ need not be true. More importantly, the functional $J(V,\pi,L)$ is not convex in $V$, $\pi$, and $L$ and there is no guarantee that a local minimum of $J(V,\pi,L)$ will be the optimal policy $\pi^*$. This is a significant difference relative to policy evaluation problems \cite{pkgtd}.
In the next section, we describe a method for solving \eqref{eq:lossprob} in parallel and then composing sparse solutions. 

\section{Composable Learning in RKHS}

To enable efficient learning in diverse environments in parallel, we require a composable representation of the control policy $\pi(\bbs)$, which is obtained through a sparse kernel parametrization. First, we develop an algorithm for learning these policy representations using a sequence of observations from a single environment. 
\subsection{Q-Learning with Kernel Normalized Advantage Functions}
To solve \eqref{eq:lossprob}, we follow the semi-gradient TD approach described by \cite{sutton2018reinforcement}, which uses the directional derivative of the loss along the direction where $\hat{y}_t(\bbs,\bba)$ is fixed and independent of $Q_t$. To obtain the semi-gradient at the sample $(\bbs_t, \bba_t, r_t, \bbs'_t)$, we define the fixed target value $y_t(\bbs_t,\bba_t,r_t,\bbs_t') = r_t + \gamma V_t(\bbs_t')$ and the temporal difference as $\delta_t = y_t - Q_t(\bbs_t, \bba_t)$. Then, we apply the chain rule to the resulting functional stochastic directional derivative, together with the reproducing property of the RKHS \eqref{eq:repprop} to obtain the stochastic functional semi-gradients of the loss $J(V,\pi,L)$ as
\begin{align}\label{eq:gradients}
\hat{\nabla}_{V} J(V,\pi,L) &= -\delta_t \kappa(\bbs_t) \;
 \\ \nonumber
\hat{\nabla}_{\pi} J(V,\pi,L)  &= - L(\bbs_t) L(\bbs_t)^T (\bba_t-\pi_t(\bbs_t)) \delta_t \kappa(\bbs_t) 
 \\
\hat{\nabla}_{L} J(V,\pi,L) &=   L(\bbs_t)^T (\bba_t-\pi_t(\bbs_t))(\bba_t-\pi_t(\bbs_t))^T \delta_t \kappa(\bbs_t)
\nonumber
\end{align}
As a result, $V, \pi, L \in \ccalH$ are expansions of kernel evaluations only at past observed states and the optimal $V$, $\pi$ and $L$ functions in the Reproducing Kernel Hilbert Space (RKHS) take the following form
\begin{align}  \label{eq:qrkhs} 
V(\mathbf{s}) &= \sum_{n=1}^N w_{V n} \kappa(\bbs_n,\bbs),  
\\
\pi(\mathbf{s}) &= \sum_{n=1}^N \bbw_{\pi n} \kappa(\bbs_n,\bbs),  \;\;\;\; 
\nonumber 
L(\mathbf{s}) = \sum_{n=1}^N \bbw_{L n} \kappa(\bbs_n,\bbs) 
\end{align}

We propose the Kernel Normalized Advantage Functions (KNAF) variant of Q-Learning to iteratively learn the action-value function while following trajectories with a stochastic policy. This algorithm includes the approximation of a kernel density metric $\rho(\bbs)$, which is later used to compose multiple learned models \eqref{eq:rhoupdate}. 
The $V$, $\pi$, $L$ and $\rho$ function representations are compressed with Kernel Orthogonal Matching Pursuit (KOMP), where we tie the compression to the learning rate to preserve tractability and convergence of the learning process \cite{mallat1993matching,polk}. 

\begin{algorithm}[H]
 \caption{Q-Learning with Kernel Normalized Advantage Functions (KNAF) }
\begin{algorithmic}[1]
 \renewcommand{\algorithmicrequire}{\textbf{Input:}}
 \renewcommand{\algorithmicensure}{\textbf{Output:}}
 \REQUIRE $l_0$, $\{ \alpha_t, \beta_t, \zeta_t, \epsilon_t, \Sigma_t \}_{t=0,1,2\ldots}$
  \STATE $V_0(\cdot) = 0, \pi_0 (\cdot) = 0, L_0 (\cdot) = l_0 I, \rho_0 (\cdot) = 0$
  \FOR {$t = 0,1,2,\ldots$}
  \STATE Obtain trajectory realization $(\bbs_t,\bba_t,r_t,\bbs_t')$ using a stochastic policy  $\pi_t(\bbs_t) \sim \mathcal{N}(\pi_t(\bbs_t),\Sigma_t) $
\STATE Compute the target value and Bellman error \\
$\hat{y}_t(\bbs_t,\bba_t,r_t,\bbs_t') = r_t + \gamma V_t(\bbs_t'), \;\;\; \delta_t = \hat{y}_t - Q_t(\bbs_t, \bba_t) $
\STATE Compute the stochastic estimates of the gradients of $J$ with respect to $V$, $\pi$ and $L$ \\
$\begin{aligned}
& \hat{\nabla}_{V} J(Q_t) = - \delta_t \kappa(\bbs_t) \\
& \hat{\nabla}_{\pi} J(Q_t) = -L(\bbs_t) L(\bbs_t)^T (\bba_t-\pi_t(\bbs_t))\delta_t \kappa(\bbs_t) \\
& \hat{\nabla}_{L} J(Q_t) =  L(\bbs_t)^T \! (\bba_t -\pi_t(\bbs_t))(\bba_t -  \pi_t(\bbs_t))^T  \delta_t \kappa(\bbs_t) 
\end{aligned}$ 
\vspace{1.0mm}
\STATE Update $V$, $\pi$, $L$, $\rho$: \\
$\begin{aligned}
& V_{t+1} = V_t(\cdot) - \alpha_t \hat{\nabla}_{V} J(Q_t), \\
& \pi_{t+1} = \pi_t(\cdot) - \beta_t \hat{\nabla}_{\pi} J(Q_t), \\
& L_{t+1} = L_t(\cdot) - \zeta_t \hat{\nabla}_{L} J(Q_t) \\
& \rho_{t+1} = \rho_t(\cdot) + \kappa(\bbs_t) 
\end{aligned}$ 
\STATE Obtain greedy compression of $V_{t+1}$, $\pi_{t+1}$, $L_{t+1}$,  $\rho_{t+1}$ via KOMP with budget $\epsilon_t$
  \ENDFOR
 \RETURN $V$,$\pi$,$L$ 
 \end{algorithmic} 
  \label{alg:knaf}
 \end{algorithm}
 
 Alg. \ref{alg:knaf} produces locally optimal policy representations of low complexity, sparsified using KOMP. These parsimonious representations reduce the complexity of real-time policy evaluation and enable the efficient policy composition procedure described in the next section.
 
  
 \subsection{Model Composition}

The key novelty of our approach is the ability to compose multiple control policies without additional training samples. We summarize a procedure for projecting $N$ policies  trained in parallel on different environments using Algorithm \ref{alg:knaf} $\pi_i(\bbs) = \sum_j w_{ij} \kappa(\bbs,\bbs_{ij})$,for $i=1,\ldots, N$, to a single function $\Pi(\bbs) = \sum_j w_{j} \kappa(\bbs,\bbs_j)$. 

We seek an interpolation between multiple candidate policies, which preserves the values of the original functions. A simple linear combination or sum is insufficient because we want to set $\Pi(\bbs) = \pi_i(\bbs)$ for all $\bbs$ in the kernel dictionaries of the candidate policies. To project multiple functions $\pi_i$ into one RKHS, we iterate through all dictionary points and perform the following update on $\Pi(\cdot)$ at $\bbs_{ij}$ given $\pi_i(\bbs_{ij})$:
\begin{equation}
\Pi(\cdot) = \Pi(\cdot) +  (\pi_i(\bbs_{ij})- \Pi(\bbs_{ij})) \kappa(\bbs_{ij},\cdot) 
\end{equation}
The next challenge we address is resolving local conflicts among $\pi_i$. We score the reliability of the $\pi_i$ in the neighborhood of $\bbs$ by the number of gradient steps performed in that neighborhood during training, which is the same as the number of observations at that state in Alg. \ref{alg:knaf}. We cannot directly use the density of kernel dictionary elements because the pruning step of Alg. \ref{alg:knaf} limits the density of the representation via compression with budget $\epsilon_t$. To accurately represent the number of observations in a neighborhood of the state space, we propose to augment the approximation of $V$, $\pi$ and $L$ with a fourth function, $\rho$, which is the kernel mean embedding of the observed states, representing the probability distribution of states observed during training \cite{sheather1991reliable}. To incorporate this approach into Algorithm \ref{alg:knaf}, we augment Step 6 of Algorithm \ref{alg:knaf} with a density estimation step:  
\begin{equation} \label{eq:rhoupdate}
\rho_{t+1} = \rho_t(\cdot) + \kappa(\bbs_t,\cdot)
\end{equation}
When $\rho$ is repeatedly pruned along with $V$, $\pi$, and $L$, the removed weights are projected onto nearby dictionary elements, producing non-unity weights for each dictionary element. This results in a sparse representation of the kernel mean embedding, accurate to $\epsilon$ over the entire state-space. This approach is inspired by kernel density estimation \cite{muandet2016kernel}.
 
 In Algorithm \ref{alg:composition}, we first accumulate dictionary points from all candidate policies into one matrix $D$ and initialize the composite policy to zero, $\Pi(\cdot)=0$. Then, we choose points $\bbs_{ij}$ from $D$ uniformly without replacement and compare the kernel density at $\bbs_{ij}$ of $\pi_i$, the function of origin for that observation, against the density at $\bbs_{ij}$ of $\pi_k$ for $k\neq i$, the other candidate functions. If the density of $\pi_i$ is greater, we add that point to the composite result because it is deemed most reliable in the neighborhood of $\bbs_{ij}$ in the state space.

\begin{algorithm}[htb]
 \caption{Composition with Conflict Resolution}
\begin{algorithmic}[1]
 \renewcommand{\algorithmicrequire}{\textbf{Input:}}
 \renewcommand{\algorithmicensure}{\textbf{Output:}}
 \REQUIRE  $\{\pi_i(\bbs) = \sum_j^{M_i} w_{ij} \kappa(\bbs,\bbs_{ij})$, \\ \hspace{6mm} $\rho_i(\bbs) = \sum_j^{M_i} v_{ij} \kappa(\bbs,\bbs_{ij}) \}_{i=1,2\ldots, N}$, $\epsilon$
 \STATE Initialize $\Pi(\cdot) = 0$, append centers $D = \lbrack \bbs_{11}, \ldots, \bbs_{ij}, \ldots \rbrack $
  \FOR {each $\bbs_{ij}\in D$ chosen uniformly at random}
  \IF{$\rho_i(\bbs_{ij}) > \max_{k \neq i} \rho_k(\bbs_{ij})$}
  \STATE  $\Pi = \Pi(\cdot) +  (\pi_i(\bbs_{ij})- \Pi(\bbs_{ij})) \kappa(\bbs_{ij},\cdot) $
  \ENDIF
  \ENDFOR
  \STATE Obtain compression of $\pi$ using KOMP with $\epsilon$
 \RETURN $f$ 
 \end{algorithmic} 
  \label{alg:composition}
 \end{algorithm}
 
For our application, we are able to estimate $\rho$ indirectly by the 
density of dictionary points of a policy $\pi$ around state $\bbs$:
\begin{equation}
\tilde{\rho}(\pi,\bbs) = \sum_{\bbs_k \in \pi} \kappa(\bbs,\bbs_{k})
\end{equation}
This simplification can be used because the structure of the reward function induces an a approximation of a kernel mean embedding on the kernel dictionary of $\pi$ itself. This does not hold for all problems. For example, if the optimal value, policy, and L functions are all zero in a region of the state space, then all kernel points in the region will be pruned away and the density of the dictionary will not accurately represent the reliability of the function.

 
 \section{Simulated \& Experimental Evaluations} 

We apply the KNAF algorithm to the robotic obstacle avoidance task by training on a variety of environments in simulation, and then validating the learned policies on a physical robot. 
\subsection{ Simulation Results}
Building on the infrastructure designed by \cite{zamora2016extending}, we train our algorithm on a wheeled robot traveling through indoor environments. The robot receives laser scans at a rate of 10 Hz, with 5 range readings at at an angular interval of $34^\circ$ with a field of view of $170^\circ$. The robot controls its angular velocity between $\lbrack -0.3,0.3 \rbrack $ rad/s at a rate of 10Hz while traveling with a constant forward velocity $0.15$ m/s. While traveling through the environment, the robot receives a reward of $-200$ for colliding with obstacles and a reward of $+1$ otherwise. The four training environments pictured in Fig. \ref{fig:diagram} are Maze, Circuit 1, Circuit 2 and Round, which mimic the appearance of indoor spaces. The Round environment is simplest of the four environments because of its radial symmetry. The Circuit 2 environment incorporates narrow hallways that turn both left and right. The Circuit 1 environment adds multiple tight turns in quick succession. The Maze environment incorporates all of these features in a more complex environment. 

\begin{figure*}
\setcounter{subfigure}{0}
\begin{subfigure}{.66\columnwidth}
\includegraphics[width=1.0\linewidth]
                {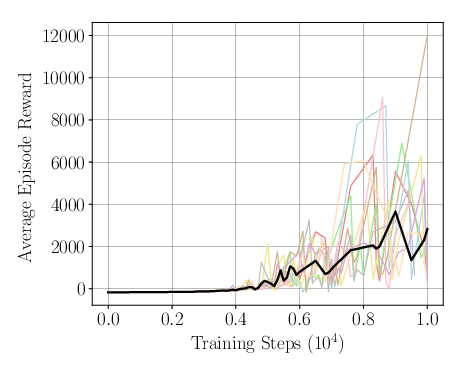}
\caption{Average Reward}
\label{subfig:avgreward}
\end{subfigure}
\hfill
\begin{subfigure}{.66\columnwidth}
\includegraphics[width=1.0\linewidth]
                {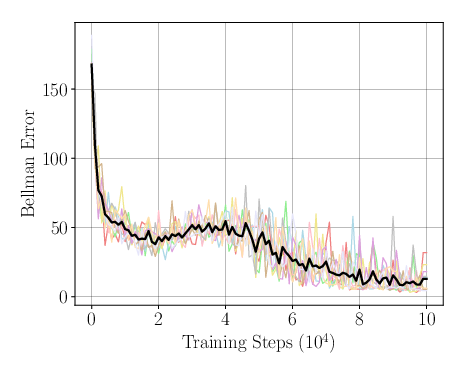}
\caption{Average Bellman Error}
\label{subfig:normbellerr}
\end{subfigure}
\hfill
\begin{subfigure}{.66\columnwidth}
\includegraphics[width=1.0\linewidth]
                {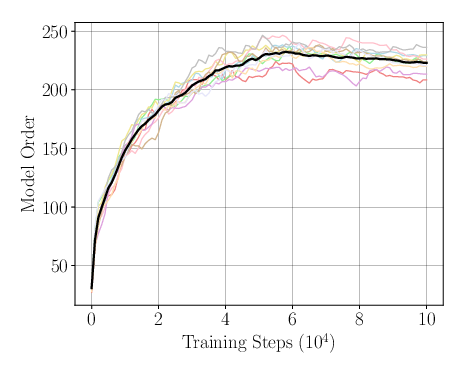}
\caption{Model Order of $Q$}
\label{subfig:modelorder}
\end{subfigure}
\caption{ 
Results of $10$ experiments over $100,000$ training steps were averaged (black curve) to demonstrate the learning progress for the robotic obstacle avoidance task with the Round environment. 
Fig. \ref{subfig:avgreward} shows the average reward obtained by the stochastic policy during training shows that the robot was able to complete 5000 simulation steps without crashing by the end of training.  Fig. \ref{subfig:normbellerr} shows the Bellman error for training samples converges to a small non-zero value. The model order of the $Q$ approximation remains under 200 for all ten experiments. 
The exploration variance $\Sigma$ was $0.2$. Constant learning rates were used, $\alpha_t = 0.25$, $\beta_t = 0.25$, $\zeta_t = 0.001$. L was initialized to $L_0 = 0.01 I$. For the KOMP Parameters, we used a Gaussian kernel with a bandwidth  of $[0.75,0.75,0.75,0.75,0.75]$ and a pruning tolerance of $\epsilon_t =  3.0$. 
}
\label{fig:roundresults}
\end{figure*}

\medskip\noindent{\bf Individual environment training.} We demonstrate the typical learning progress of Algorithm \ref{alg:knaf} by reproducing ten trials of training with the Round environment, and plot training rewards, model order, and Bellman error in Fig. \ref{fig:roundresults}. The learning progress is extremely reproducible, with average episode rewards of at least 2000 after 100,000 training steps. By the end of training, the robot travels for at least 2000 steps before crashing during every episode. Training Bellman error reliably converges for every trial. The resulting policies are parsimonious, with a limiting model order of fewer than 250 kernel dictionary elements. 
Next, we demonstrate the performance of Alg \ref{alg:knaf} on the four environments pictured in Fig. \ref{fig:diagram} in Table \ref{results_table}. All policies achieve zero crashes during testing. We also observe that the Maze environment required many more training steps and achieved a larger model order. We interpret this to mean that the Maze environment is more varied, and therefore requires a higher coverage of the state space. 

\medskip\noindent{\bf Generalization across environments.} We analyze the inability of a single policy to generalize to other environments in Table \ref{leavepout}. We observe that policies trained on the Round environment (Policy 1) could not generalize to other environments at all, with an average test reward of $-4226$. We observe that all policies were able to successfully navigate the Round environment with no crashes, most likely due to its simplicity. The policy trained on the Maze environment (Policy 2) was able to somewhat generalize to the other three environments, while none of the other environments could generalize to the Maze, due to its complexity. Policies trained on Circuit 1 (Policy 4) and Circuit 2 (Policy 3) completely failed on the Maze environment, but were able to transfer somewhat among each other, and to the Round environment. 

\medskip\noindent {\bf Dual compositions.} We apply the composition algorithm (Alg. \ref{alg:composition}) to combine two of the four trained policies to observe performance on the original two environments and analyze generalization performance on the remaining environments in Fig. \ref{leavepout}. All controllers composed from Policy 2 (trained on Maze) achieved perfect reward of 1000 on the difficult Maze environment. Policies composed from Circuit 1 and Circuit 2 were able to achieve positive average rewards on the Circuit 1 and 2 environments. We observe some generalization capabilities in the results for the composite Policy 1/2 (Round and Maze), which performs better in all four test environments than both of the original policies. The other dual composite policies were not able to generalize better than the individual policies. 

\medskip\noindent{\bf Composition of multiple policies.} We further validate the composition algorithm (Alg. \ref{alg:composition}) using every combination of four trained policies: an additional eleven policies composed from two, three, and four of the original four policies. The performance of the composite policies was validated on the four training environments in Fig. \ref{leavepout}. Nearly optimal performance was observed when all four policies (1/2/3/4) were composed into one, and then tested on all of the four environments. Two collisions were observed in the Circuit 1 environment, and none in the other three when testing for 1,000 time steps in each environment. 

 \subsection{Robot Results} 

\begin{figure} 
\includegraphics[width=0.97\linewidth]
                {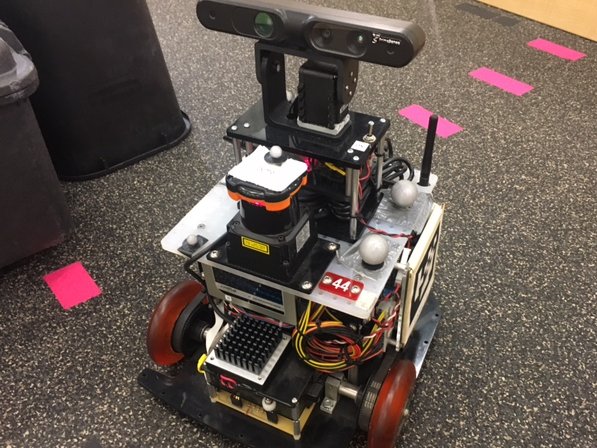}
\caption{ 
Scarab robot with equipped with an  on-board  computer, wireless communication, and a Hokuyo URG laser range finder. 
}
\label{fig:robot}
\end{figure}

\begin{figure} 
\includegraphics[width=0.97\linewidth]
                {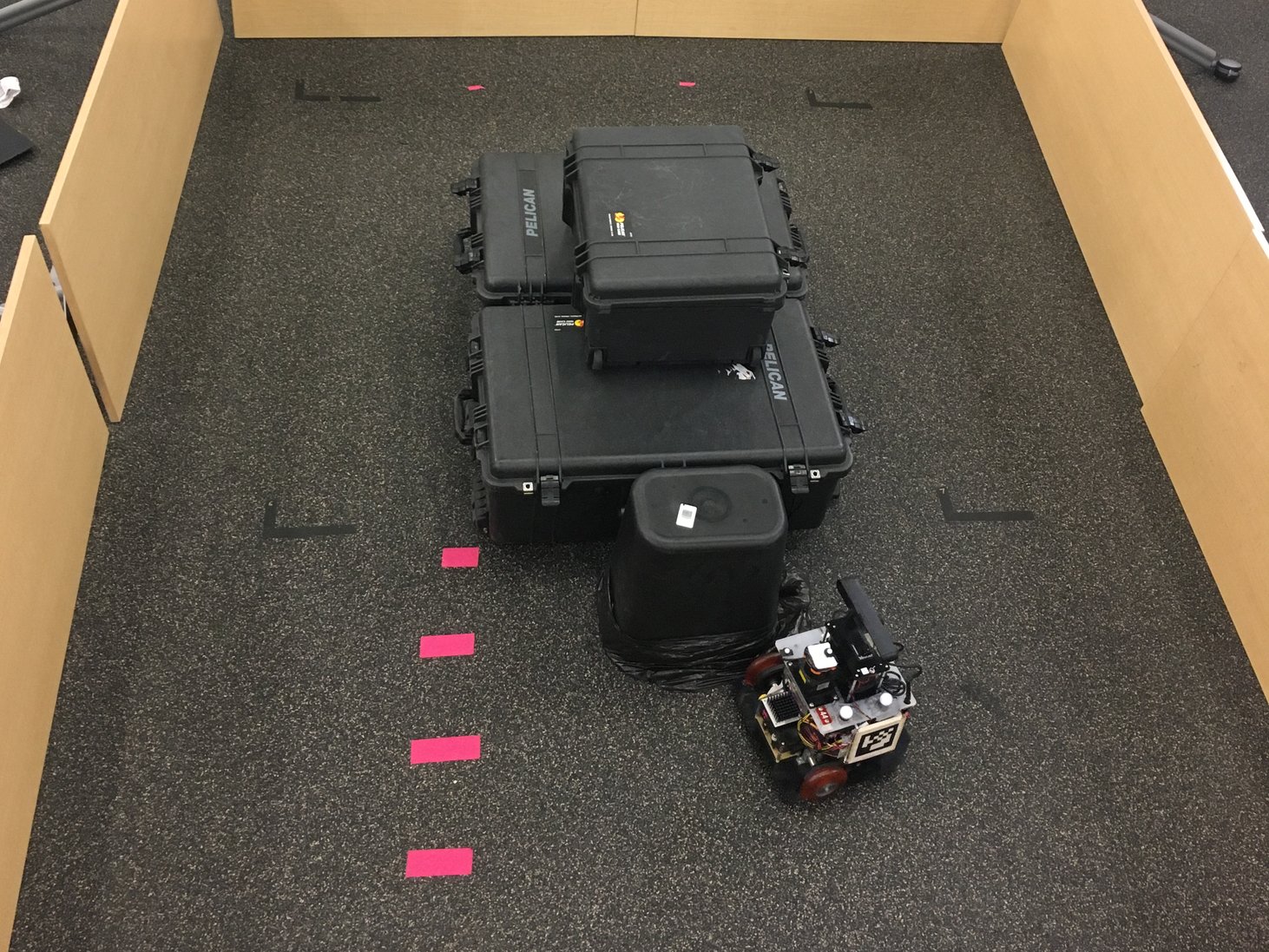}
\caption{ We validate our approach by testing policies trained in simulation on a real robot in a laboratory environment. The policy trained only on the Round environment experienced 3 crashes over 1,000 testing steps. The composite 1/2/3/4 policy in Table \ref{leavepout}, combined from all four policies, received a reward of 1,000 during 1,000 testing steps, with no crashes.  }
\label{fig:robotenv}
\end{figure}

\begin{figure} 
\includegraphics[width=0.93\linewidth]
                {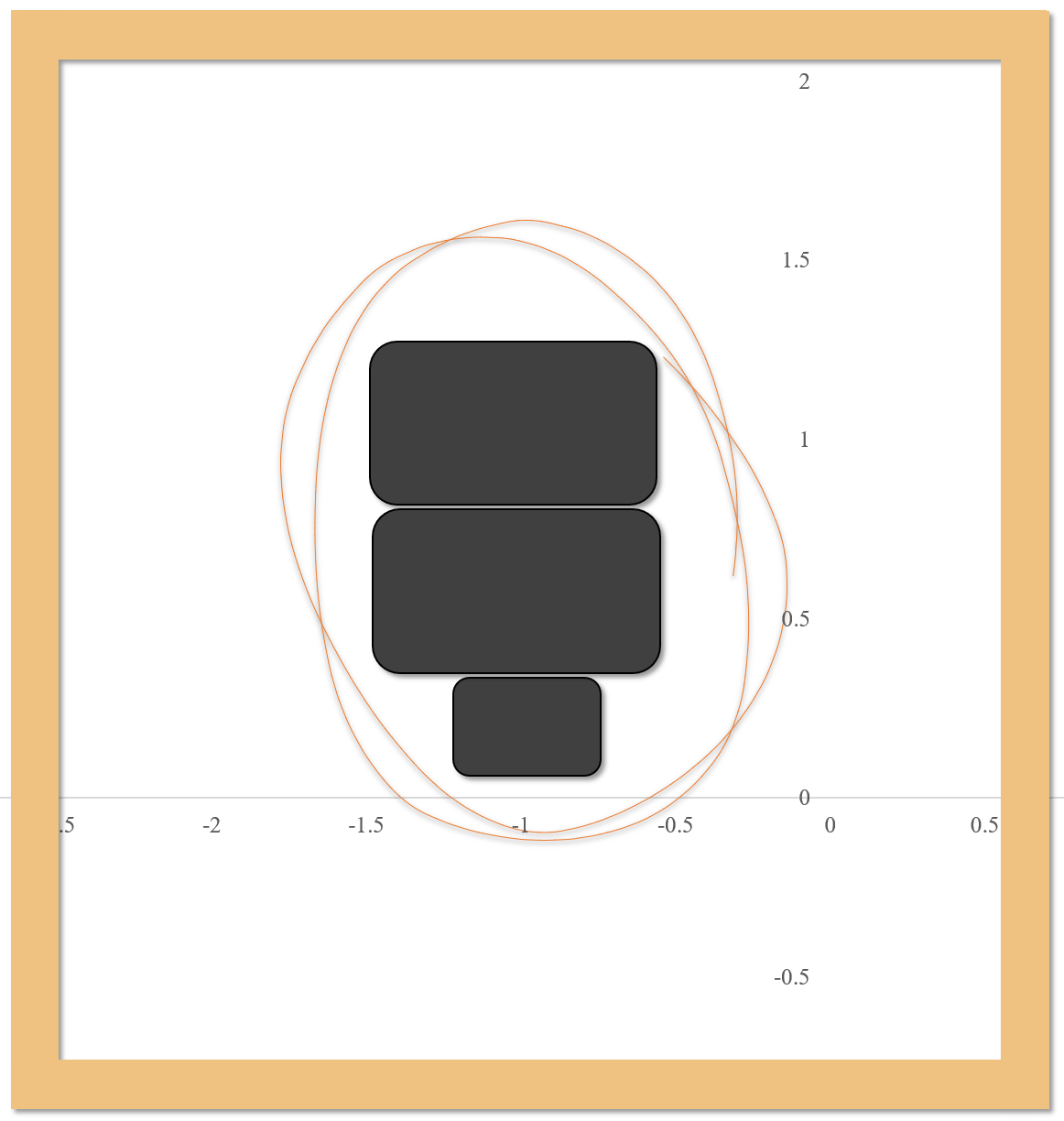}
\caption{ We visualize the trajectory observed when testing the composite policy (Policy 1/2/3/4) in Table \ref{leavepout} on the Scarab robot. Position coordinates were computed through onboard motor odometry, which results in a small drift in the position. }
\label{fig:robotpath}
\end{figure}

The  real-world  validation experiments  were  carried  out on a Scarab robot pictured in Fig. \ref{fig:robot}, equipped with an  on-board  computer, wireless communication, and a Hokuyo URG laser range finder. It is actuated by stepper motors and its physical dimensions are 30 x 28 x 20 cm with a mass of 8kg \cite{atanasov2012stochastic}. Laser scans were received at a rate of 10Hz and 5 range readings were obtained as in simulation. Angular velocity commands were issued at 10Hz. The test environment was built using laboratory furniture and miscellaneous equipment pictured in Fig. \ref{fig:robotenv}. When testing the policy trained only on the Round environment, we observed 3 collisions, with a total reward of 397 over 1,000 time steps. Using the composite policy, which incorporates all 4 policies, no collisions were observed during 1,000 time steps. 
In Fig. \ref{fig:robotpath}, we visualize the trajectory obtained by testing the composite policy on the Scarab robot, which shows that the robot successfully completed multiple loops around the obstacles in the environment. 

\section{Conclusions \& Future Work}

We envision multi-robot systems which can exchange concise representations of their knowledge to enable distributed learning and control in complex environments with limited communication \cite{schlotfeldt2018anytime}. In this work, we take the first step towards this goal by developing and demonstrating the Q-Learning with KNAF algorithm, which allows the composition of multiple learned policies without additional training samples. Rather than a single robot collecting information about multiple environments in sequence, multiple robots could work in parallel and combine their models. We also see a clear path to extend this approach to higher-dimensional problems using an auto-encoder to learn policies in an RKHS based on a feature space representation, rather than directly in the state space \cite{richter2017safe}. 

 
\begin{table}
\centering
\begin{tabular}{l l l l l}
\Xhline{2\arrayrulewidth}
Environment & Steps & Model Order & Loss & Rewards \\
\hline 
Round       & 230K  & 224 & 2.24 & \textbf{1000}    \\
Maze        & 630K     & 779           & 16.04             & \textbf{1000}                \\
Circuit 2   & 280K     & 467           & 36.40              & \textbf{1000}                \\
Circuit 1   & 500K     & 578 & 47.97 &  \textbf{1000}                \\
\hline
\end{tabular}
\caption{ 
Limiting model complexity, training loss (Bellman error), and accumulation of rewards over 1,000 testing steps for each of the four environments. All environments are solved by the K-NAF algorithm within 700,000 simulation steps.
}\label{results_table}
\end{table}

 
\begin{table}
\centering
\begin{tabular}{lllll}
\Xhline{2\arrayrulewidth}
Policies / Reward      & Round & Maze & Circuit 2 & Circuit 1 \\
\hline 
1 - Round     & \textbf{\textcolor{ForestGreen}{1000}} & \textcolor{BrickRed}{-11663}  & \textcolor{BrickRed}{-608}  & \textcolor{BrickRed}{-608} \\
2 - Maze      & \textbf{\textcolor{ForestGreen}{1000}} & \textbf{\textcolor{ForestGreen}{1000}} & \textcolor{BrickRed}{-5}  & \textcolor{BrickRed}{-407} \\
3 - Circuit 2 & \textbf{\textcolor{ForestGreen}{1000}} & \textcolor{BrickRed}{-11663}        & \textbf{\textcolor{ForestGreen}{1000}}                & \textcolor{ForestGreen}{196}              \\
4 - Circuit 1  & \textbf{\textcolor{ForestGreen}{1000}} & \textcolor{BrickRed}{-11462}         & \textcolor{BrickRed}{-407}               & \textbf{\textcolor{ForestGreen}{1000}}               \\
1 / 2          & \textbf{\textcolor{ForestGreen}{1000}}           & \textbf{\textcolor{ForestGreen}{1000}}        & \textcolor{BrickRed}{-5}                  & \textcolor{BrickRed}{-206}             \\
1 / 3          & \textbf{\textcolor{ForestGreen}{1000}}           & \textcolor{BrickRed}{-11663}        & \textcolor{ForestGreen}{799}                 & \textcolor{BrickRed}{-206}              \\
1 / 4          & \textbf{\textcolor{ForestGreen}{1000}}           & \textcolor{BrickRed}{-11261}      & \textcolor{BrickRed}{-206}               & \textcolor{ForestGreen}{799}               \\
2 / 3           & \textbf{\textcolor{ForestGreen}{1000}}          & \textbf{\textcolor{ForestGreen}{1000}}         & \textbf{\textcolor{ForestGreen}{1000}}               & \textcolor{BrickRed}{-5}                  \\
2 / 4           & \textbf{\textcolor{ForestGreen}{1000}}          & \textbf{\textcolor{ForestGreen}{1000}}          & \textcolor{BrickRed}{-5}                 & \textcolor{ForestGreen}{799}                \\
3 / 4           & \textbf{\textcolor{ForestGreen}{1000}}           & \textcolor{BrickRed}{-11462}         & \textcolor{ForestGreen}{397}                  & \textcolor{ForestGreen}{397}              \\
1 / 2 / 3         & \textbf{\textcolor{ForestGreen}{1000}}           & \textbf{\textcolor{ForestGreen}{1000}}          & \textcolor{ForestGreen}{799}                     & \textcolor{ForestGreen}{196}                  \\
1 / 2 / 4         & \textbf{\textcolor{ForestGreen}{1000}}           & \textbf{\textcolor{ForestGreen}{1000}}            & \textcolor{BrickRed}{-5}                   & \textbf{\textcolor{ForestGreen}{1000}}               \\
1 / 3 / 4         & \textbf{\textcolor{ForestGreen}{1000}}           & \textcolor{BrickRed}{-11663}        & \textcolor{ForestGreen}{397}                & \textcolor{ForestGreen}{799}               \\
2 / 3 / 4         & \textbf{\textcolor{ForestGreen}{1000}}          & \textbf{\textcolor{ForestGreen}{1000}}          & \textcolor{ForestGreen}{799}                 & \textcolor{BrickRed}{-206}              \\
\textbf{1 / 2 / 3 / 4}       & \textbf{\textcolor{ForestGreen}{1000}}              & \textbf{\textcolor{ForestGreen}{1000}}            & \textbf{\textcolor{ForestGreen}{1000}}                 & \textcolor{ForestGreen}{598}     \\
\hline
\end{tabular}
\caption{
Cross-validation was performed on all possible compositions of the original 4 policies, each of which was trained on one environment pictured in Fig. \ref{fig:diagram}. Each row is a policy composed using Alg \ref{alg:composition} of one or more policies trained using Algorithm \ref{alg:knaf}. All 15 policies were tested on the 4 environments, and compared based on the reward accumulated over 1,000 time steps in each environment. 
}\label{leavepout}
\end{table}


\section*{ACKNOWLEDGMENT}
We thank Arbaaz Khan, Brent Schlotfeldt, Dinesh Thakur, Sikang Liu, and Ty Nguyen for their help with brainstorming and experimental frameworks. 

 
\bibliographystyle{IEEEtran}
\bibliography{IEEEfull,bibliography}

\end{document}